\documentclass[runningheads]{llncs}
\def\BibTeX{{\rm B\kern-.05em{\sc i\kern-.025em b}\kern-.08em
   T\kern-.1667em\lower.7ex\hbox{E}\kern-.125emX}}
\usepackage{graphicx}
\usepackage{mathtools}
\usepackage[utf8]{inputenc}
\usepackage{lipsum} 
\usepackage{booktabs}
\usepackage[ruled]{algorithm2e}
\usepackage{url}
\usepackage{colortbl}
\usepackage{multicol}
\usepackage{color}
\usepackage{multirow}
\usepackage{amsfonts} 
\usepackage{enumitem}
\usepackage{array}
\usepackage{siunitx}
\usepackage[labelfont={bf}]{caption}
\usepackage{balance}
\usepackage{graphicx}
\usepackage{array} 
\usepackage{todonotes}
\usepackage{subfig}
\usepackage[nolist,nohyperlinks]{acronym}
\usepackage{footnote}
\usepackage[perpage]{footmisc}
\usepackage[autostyle]{csquotes}
\makesavenoteenv{tabular}
\usepackage{hyperref}

\author{Zheyuan Zhang \and
Ulas Bagci}
\institute{Department of Biomedical Engineering, Northwestern University, IL 60201, USA \\
\email{ulas.bagci@northwestern.edu}}
\authorrunning{Zhang et al.}
\begin{document}
\title{Dynamic Linear Transformer for 3D Biomedical Image Segmentation}
\maketitle              
\begin{abstract}
Transformer-based neural networks have surpassed promising performance on many biomedical image segmentation tasks due to a better global information modeling from the self-attention mechanism. However, most methods are still designed for 2D medical images while ignoring the essential 3D volume information. The main challenge for 3D Transformer-based segmentation methods is the quadratic complexity introduced by the self-attention mechanism \cite{vaswani2017attention}. In this paper, we are addressing these two research gaps, lack of 3D methods and computational complexity in Transformers, by proposing a novel Transformer architecture that has an encoder-decoder style architecture with linear complexity. Furthermore, we newly introduce a dynamic token concept to further reduce the token numbers for self-attention calculation. Taking advantage of the global information modeling, we provide uncertainty maps from different hierarchy stages. We evaluate this method on multiple challenging CT pancreas segmentation datasets. Our promising results show that our novel 3D Transformer-based segmentor could provide promising highly feasible segmentation performance and accurate uncertainty quantification using single annotation. Code is available \url{https://github.com/freshman97/LinTransUNet}.
\keywords{Linear Transformer  \and Pancreas Segmentation\and Uncertainty Quantification.}
\end{abstract}
\section{Introduction}

Long-range dependence is needed and essential for the challenging segmentation problems in biomedical images. Transformer is one novel architecture that allows us to capture such information and has achieved promising performance on many vision  tasks including biomedical image segmentation \cite{vaswani2017attention,cao2021swinunet,huang2021missformer}. A large amount of Transformer-based biomedical image segmentation methods has been proposed in the last two years. For example, the Swin-Unet replaces the traditional convolution blocking with the Transformer as the feature extraction and reconstruction tools in the encoder and decoder process \cite{cao2021swinunet}. However, it is still challenging to apply such architecture to 3D images due to the huge computation challenges. The traditional self-attention mechanism requires \textit{O}($N^2$) complexity, where N represents the feature length. This corresponds to \textit{O}($(hwd)^2$) complexity, where h, w, d represent the height, width, depth of volume respectively. This is beyond the maximum memory for most current GPUs. This limits the Transformer's application in 3D segmentation for biomedical images. To address this significant burden, some methods have been proposed to reduce the number of tokens for 3D segmentation. For instance, the TransUnet inserted the Transformer block only after the deepest layer of feature extraction in the traditional Unet segmentation structure to model the long-distance dependence \cite{chen2021transunet}. UNETR applied the self-attention after dividing into a sequence of uniform
non-overlapping large patches, thus reducing the number of feature lengths \cite{hatamizadeh2022unetr,tang2022self}. However, it is obvious that these segmentation method limits global information modeling only in the higher level feature layer, not otherwise.

In this paper, we propose a new dynamic linear Transformer algorithm for 3D biomedical image segmentation enabling us to apply the Transformer on volumetric data with \textit{linear complexity}. Recently linear computation self-attention could reduce the complexity from \textit{O}($N^2$) to \textit{O}($N$) which dramatically reduced the computation requirement \cite{kitaev2020reformer,wang2020linformer,shen2021efficient}. Furthermore, the segmentation target is located within a certain physical region rather than the whole frame for many biomedical image segmentation tasks. Thus, almost all previous works used a two-step segmentation strategy to improve the segmentation performance \cite{zhao2019twostage}. This guided us to question that \textit{"do we really need to apply self-attention within the whole volume?"} Here, we apply a new dynamic strategy to further limit the computation within the region of interest (ROI). This strategy could further reduce the computation to $(1+\beta)^2 \alpha^2$ where $\alpha$ is the relative ratio of ROI size over the whole frame and $\beta$ is the extension ratio. Besides, by combining the global information the hierarchy decoder structure could simultaneously provide one pixel-level uncertainty map for segmentation.

\textbf{Clinical Applications:} The pancreas is one important abdominal organ that plays an essential role in producing enzymes for digestion and  producing insulin and glucagon to adjust blood sugar levels. The segmentation of the pancreas from radiology scans is one crucial work for pancreatic disease diagnosis and treatment \cite{roth2015deeporgan}. However, this work is very time-consuming and labor-intensive. There are many challenges that limit designing automatic segmentation algorithms for the pancreas. First, the pancreas is one small organ with a sparse and complex shape compared to other abdominal organs. Second, pancreas has one elongating structure which can be hardly defined and long-range dependence is extremely significant for this organ. Third, the contrast of the pancreas in computed tomography (CT) scans is relatively low. Here, we validate our method on two challenging pancreas segmentation datasets to prove the effectiveness of the proposed method.

\section{Methods}
\subsection{Transformer with linear complexity}
The self-attention mechanism achieved remarkable performance in many computer vision and natural language processing tasks. However, the quadratic complexity prohibits its application in volumetric segmentation tasks. Recently some  methods have been proposed to reduce the computation of self-attention to linear complexity \cite{kitaev2020reformer,wang2020linformer,shen2021efficient}. 

Given the individual feature vectors $X \in \mathbb{R}^{n\times d}$, we have the values $V \in \mathbb{R}^{n\times d}$, the queries $Q \in \mathbb{R}^{n\times d}$, the keys $K \in \mathbb{R}^{n\times d}$ after linear projections, where n represents the feature length and d represents the feature dimension. The traditional self-attention mechanism can be expressed in a more general way as follow:
\begin{equation}
    V_{i}^{'} = \sum_{j=1}^{n} sim(Q_i, K_j) V_j
\end{equation}
where the similarity function is defined as $sim(q,k) = softmax(qk^T/\sqrt{d})$ and i, j are the index for features. In the linear Transformer we want to replace the similarity function such that the similarity function can be divided into two separate parts using the normalized feature quantification function $sim(q, k) = \phi(q) \rho(k)^T$. This replacement could allow us get:
\begin{equation}
    V_{i}^{'} = \sum_{j=1}^{n} (\phi(Q_i) \rho(K_j)^T) V_j= \phi(Q_i)( \sum_{j=1}^{n} \rho(K_j)^T V_j),
\end{equation}

In this paper, we employ the similarity definition of the Efficient Transformer \cite{shen2021efficient}, $\phi(Q_i), \rho(K_j)$ denote applying the softmax function along each row or column of $Q, K$. This definition allows us to keep the important property of original self-attention that $\sum_{j=1}^{n} sim(Q_i, K_j) = 1$ and reduce the complexity from quadratic to linear.
\begin{figure}
\includegraphics[width=\textwidth]{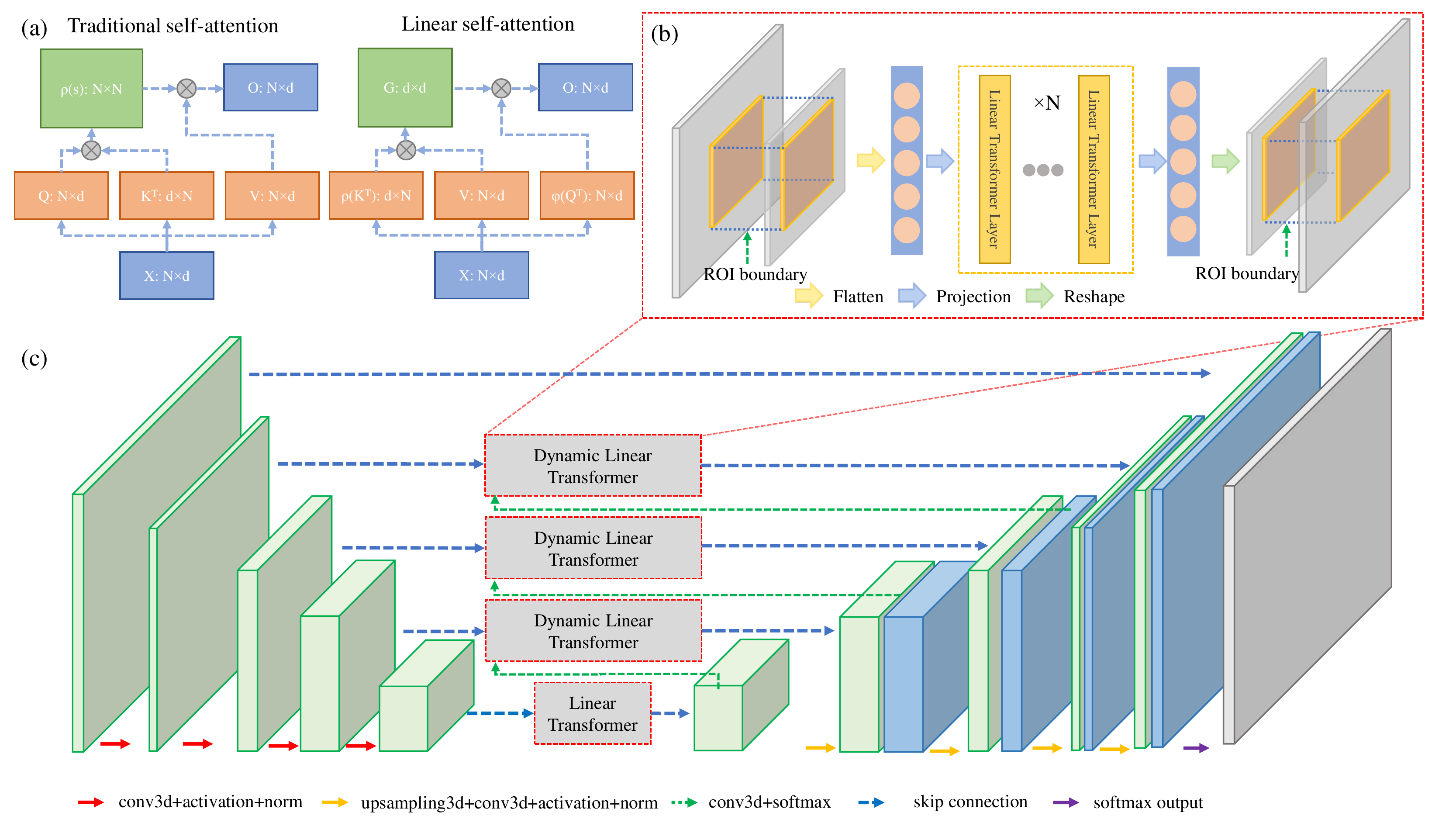}
\caption{(a) Comparison of traditional self-attention mechanism v.s. linear self-attention mechanism. (b) The dynamic Transformer block using linear self-attention. (c) The encoder-decoder style segmentation engine.} \label{fig1}
\end{figure}
\subsection{Dynamic Transformer}
Most Transformer-based image segmentation models are using all the tokens extracted from the raw image. However, we have one question that do we really need all the tokens, and are all the tokens equally meaningful for segmentation? The answer is clearly "no" as for most biomedical image segmentation tasks, the target region usually is located with a certain region due to the physical world limitation. For example, we shall not find tumors located in the kidney if we are looking for liver tumors. Thus the tokens extracted from other regions will be one waste for computation and even further introduce the false positive. In the traditional convolution neural network (CNN), two-stage segmentation methods are widely used to address this problem. Recently, a dynamic Transformer concept is proposed to automatically extract the useful tokens for each input image and reduce the computation  for image classification \cite{rao2021dynamicvit}. This inspires us to introduce the dynamic concept from classification into segmentation. We further limit the dense tokens extraction within the ROI regions while in the rest of the image we only take sparse tokens. The ROI region will be decided by the output mask from the previous segmentation layer, which will auto make the whole attention process while keeping the computation linear. 

The implementation algorithm of this dynamic sampling is shown in Figure \ref{fig1}(b). According to the ROI region size, we determine two sampling rates: one high sampling rate for the ROI region and one low sampling rate for the rest of the region dynamically. For instance, given the ROI size $x_0$ in the width direction with the whole size of $w$, we further set two hyper-parameter the relative ROI ratio ($\alpha$) and extension ratio ($\beta$). Then the high sampling rate in the ROI region will be $\frac{\alpha w}{x_0}$, while the sampling rate for the rest of the region will be $\frac{\beta \alpha w}{w-x_0}$. Thus, in this direction the size of the image will be reshaped into $(1+\beta) \alpha w$ after this ROI resizing process. In this way, we reduce the token numbers for the self-attention calculation to $(1+\beta)^2 \alpha^2 $ of original token numbers. Here we only apply the dynamic ROI resizing in height and widths direction while keeping the original size in the depth direction due to relatively small size in the depth direction. The extracted token after ROI resizing is served into the linear Transformer block to model the long-distance dependence. Next, the image is resized back to its original size according to the spatial position. In this work,  we set the $\alpha=0.5$ according to the pancreas size and $\beta=0.2$, thus we could reduce the overall computation by $51\%$. Here these two parameters could be adjusted according to the potential segmentation target size, thus the most of computation is limited within the region of interest rather than the whole image.
\subsection{Network structure}
After introducing the linear self-attention and dynamic block, we build one the encoder-decoder style segmentation engineer enhanced with self-attention shown in Figure \ref{fig1}(c). As shown in the literature \cite{oktay2018attentionunet}, the encoded feature map is noisy which limits the performance of self-attention. Thus, the segmented layer also serves as one spatial attention before the ROI resampling. Beside, similar image embedding and embedding are used in the input and output layer respectively in order to reduce the image size \cite{cao2021swinunet}. Other setups are similar to the attention UNet \cite{oktay2018attentionunet}. During the encoder process, we extract the higher-level representative features. In the decoder process, the extracted features are used to generate one segmentation mask at each hierarchy level. As we could expect, the deeper level segmentation loses more detailed information while maintaining one relatively accurate positional information. This positional information can be used in the higher level skip connection to perform the dynamic Transformer block to reduce the computation. In other words, the importance of lower level features will be decided by the deeper level segmentation masks. Thus, in the final output layer, we could combine the accurate positional information coming from the deep feature level and the low-level features to provide one accurate boundary.
\section{Results}
\subsection{Datasets and Training details}
The first pancreas dataset used in the experiments is the NIH pancreas dataset which contains 82 abdominal contrast-enhanced 3D CT scans \cite{roth2015deeporgan}. These CT scans have resolutions of $512\times512$ pixels with varying pixel size and slice thickness between 1.5-2.5mm. The second dataset used in the experiments is the Medical Segmentation Decathlon which contains 282 CT scans with annotation \cite{antonelli2021medical}. The training and test dataset is split into 7:1 randomly. Traditional data augmentations including random crop, random rotate, random zoom are employed during the training. We clipped the CT value between -100 and 250 Hu(Hounsfield Unit). Then we normalized the image by extracting the mean value and dividing it by the standard deviation according to the foreground intensity distribution of each pancreas dataset. Note here we directly serve the $512 \times 512 \times 32 (Depth)$ volumetric images into the network without slicing into patch.

We implement the proposed network structure using PyTorch with 2 NVIDIA RTX A6000 GPUs. The optimizer is set as the AdamW which is widely used to train the Transformer. The learning rate is set as 0.0001 with one delay rate of 0.5 after 100 epochs and the batch size is set as 8. The training losses are set as the combination of \textit{Dice loss} and \textit{Cross-Entropy loss}, given the pixel position $j$ and the predicted possibility value  $q_{j}$, the label value $p_{j}$:
\begin{equation}
    loss =  (1-\frac{\sum_{j} p_{j} q_{j}}{\sum_{j} p_{j} + q_{j}})+ \frac{1}{n}\sum_{j=1}^{n}(p_{j} log(q_{j}) + (1-p_{j})log(1-q_{j})).
\end{equation}
Deep supervision is applied for each decoder layer to guide the training. Thus, the total loss will be the summation of the loss value at each hierarchy stage $i$,
\begin{equation}
    total\_loss =  \sum_i loss_i.
\end{equation}
\subsection{Segmentation Performance}
\begin{figure}
\includegraphics[width=\textwidth]{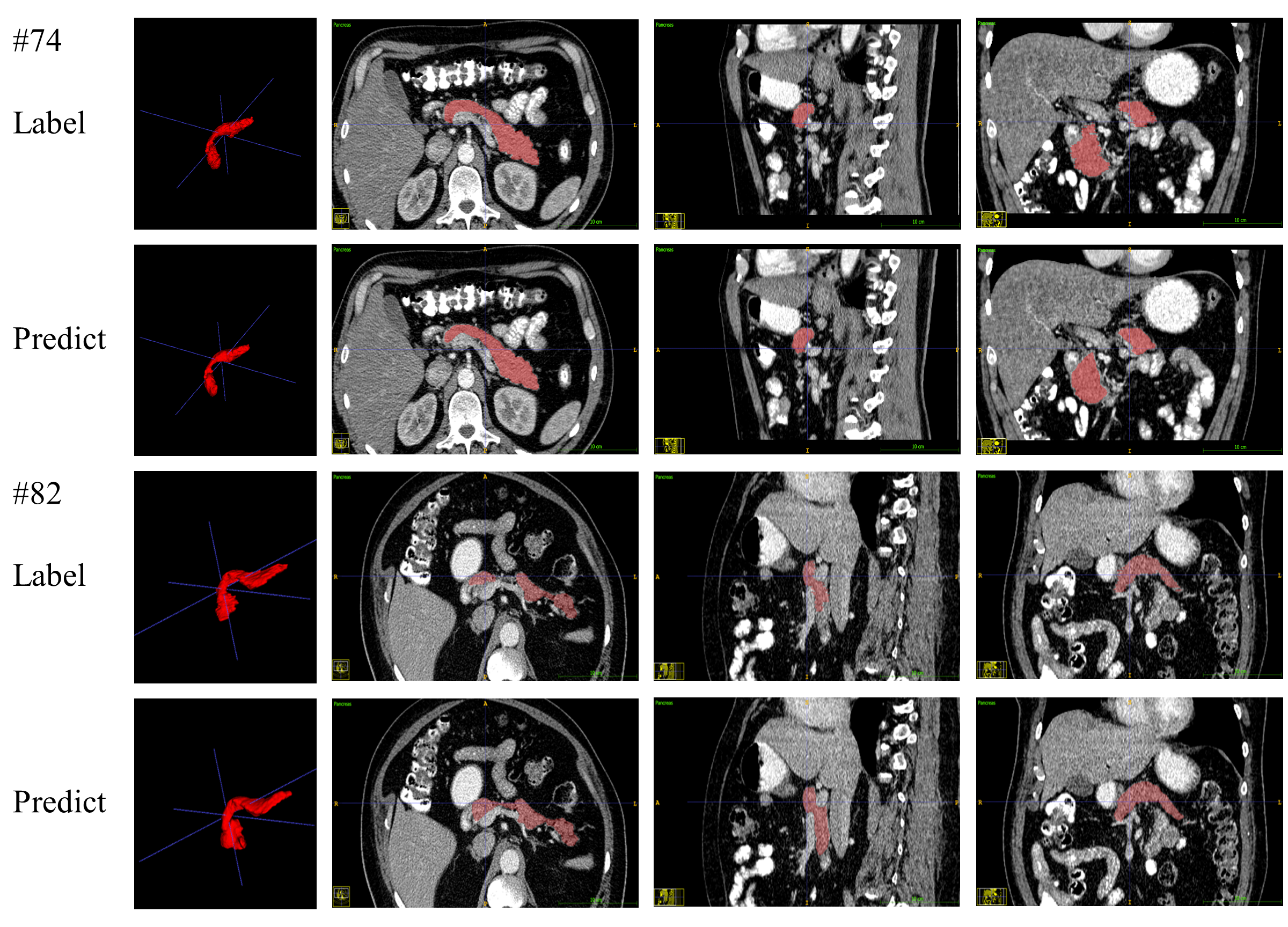}
\caption{Some pancreas segmentation examples visualization.} \label{fig2}
\end{figure}
Firstly, we trained our network from scratch on NIH pancreas data. The network converged after about 600 epochs of training. Table \ref{tab1} showed the performance of our model and the comparison with other methods. Note here all the selected models are trained from scratch without external datasets. One most straightforward comparison is the performance difference between the designed architecture and the attention UNet architecture \cite{oktay2018attentionunet}. Except for the introduction of the linear self-attention mechanism, there is no difference in structure level. We could clearly observe performance improvement by introducing long-distance modeling, which proves the significance of this structure. Figure \ref{fig2} illustrates two independent segmentation cases in the NIH dataset. The $\#74$ cases achieved one \textit{Dice coefficient} of 0.90 which is higher than the average. The $\#82$ cases achieved one \textit{Dice coefficient} of 0.80 which is lower than the average. We could observe that the proposed model provides one accurate segmentation boundary even compared with the radiologist's annotation, having consistently higher rates with smaller deviations. Particularly, Yu et al. \cite{yu2018recurrent} employed a recurrent saliency transformation network to incorporate multi-stage visual cues with coarse segmentation and fine segmentation two stages. Our model surpassed this performance without specific two-stage training. Although fewer researches provide one evaluation metric based on the surface distance, we could observe that our model achieved one mean surface distance (MSD) of 0.984mm, which indicates the effectiveness of our model.
\begin{table}
\caption{The segmentation performance on the NIH dataset, we employed the Dice coefficient (Dice), Precision, Recall and Mean Surface Distance (MSD) as the evaluation metrics}\label{tab1}
\begin{tabular}{|c|c|c|c|c|}
\hline
Methods & Dice & Precision & Recall & MSD (mm)\\
\hline
3D UNet\cite{oktay2018attentionunet} & $ 0.815 \pm 0.068 $ & $0.815 \pm 0.105$ & $0.826 \pm 0.062 $& 2.576 $\pm$ 1.180\\
Attention U-Net\cite{oktay2018attentionunet} & $ 0.821\pm 0.057$ & $0.815 \pm 0.093$ & $0.835 \pm 0.057 $&2.333 $\pm$ 0.856 \\
MDS-Net\cite{li2020MDS} & $ 0.835 \pm 0.062 $ & $\textbf{0.845} \pm 0.069$ & $0.837 \pm 0.104 $&-\\
Liu et al.\cite{liu2019automatic} & $ 0.841 \pm 0.049 $ & $0.836 \pm 0.059$ & $0.853 \pm 0.082 $&-\\
Yu et al.\cite{yu2018recurrent} & $ 0.845 \pm 0.050 $ & - & - &-\\
\hline
Ours & \textbf{0.855}$\pm$ 0.037 & 0.840 $\pm$ 0.083 & \textbf{0.882} $\pm$ 0.040 & \textbf{0.984} $\pm$ 0.399 \\
\hline
\end{tabular}
\end{table}

Secondly, we evaluate our model's performance on the Medical Segmentation Decathlon dataset. We achieved one 0.833 dice coefficient with 0.831 precision and 0.862 recall. This performance further proved that the model could achieve one relatively accurate segmentation on the challenging pancreas segmentation datasets

\subsection{Uncertainty Quantification}
Taking advantage of the long-distance modeling ability, the decoded mask at every stage could provide sufficient global information. This allows us to acquire one \textit{uncertainty distribution} over the predictions\cite{hong2021hypernet}. Given the predict mask $predict_{i}$ in different hierarchy stages $i$ and final binary output segmentation map $out$, the definition of the pixel-wise uncertainty is given as follow:
\begin{figure}
\includegraphics[width=\textwidth]{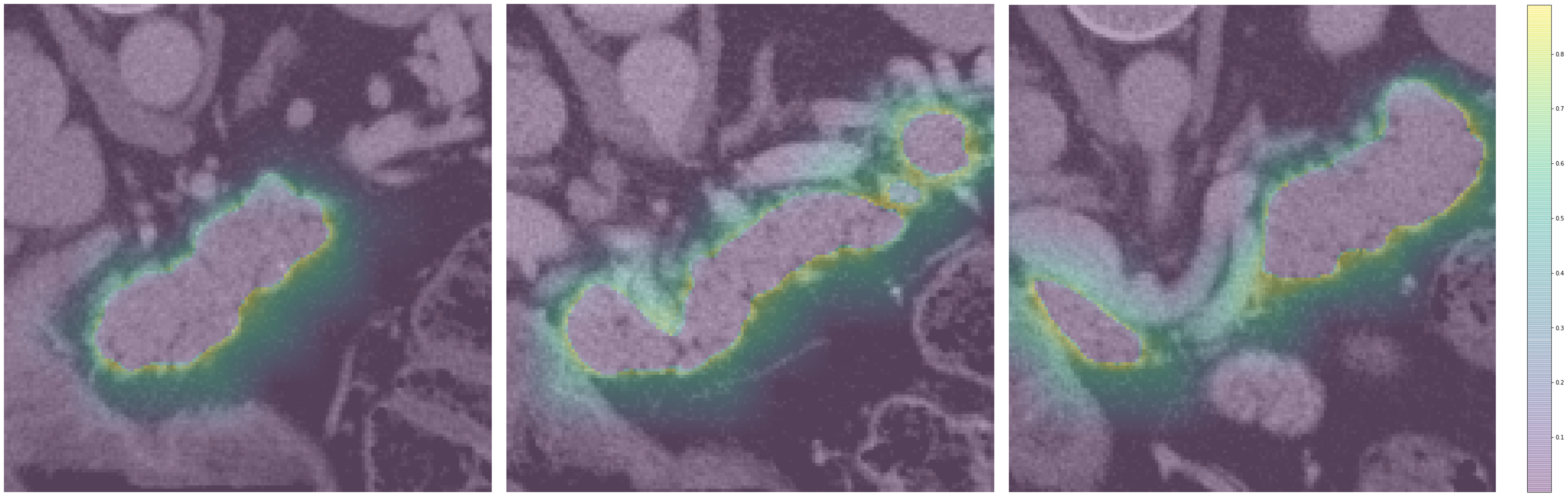}
\caption{Uncertainty map visualization, image is cropped according to the pancreas position} \label{fig3}
\end{figure}
\begin{equation}
    Uncertainty = \sqrt{\frac{1}{N}\sum_{i=1}^{N} (predict_{i} - out)^2}
\end{equation}

We anticipate that the high uncertainty value should locate around the boundary. Figure \ref{fig3} clearly demonstrates that the highest uncertainty values are located around the pancreas boundaries. Furthermore, we calculate the uncertainty value in the true-positive (TP) of $ 0.040 \pm 0.009 $, true negative (TN) of $0.002 \pm 0.000 $, false negative (FN) of $0.122 \pm 0.019 $, false positive (FP) of $0.563 \pm 0.036 $. We could clearly observe that the mispredicted region has one relatively higher uncertainty value compared with the corrected predicted region with one p value of $5.3\times10^{-9}$. This agrees with our anticipation that higher uncertainty region is more likely to be mispredicted or, not in agreement with human annotation.
%
\section{Discussion and Concluding Remarks}
Although this network structure has achieved promising performance, there are still several improvements that are needed in the future. Particularly on the NIH pancreas dataset, Salanitri et al. \cite{proietto2021hierarchical} achieved current state of the art with one Dice coefficient of 0.880 using hierarchical feature learning. It is well known that the training for self-attention-based structure requires a larger training dataset compared with the traditional CNN network \cite{xu2020optimizing}, which limits the model's performance on the small dataset like the pancreas segmentation datasets. We could clearly observe that there is one performance gap between the training (Average Dice coefficient: 0.910) and test dataset (Average Dice coefficient: 0.855). A large level pertaining on medical image or external dataset would be helpful to minimize this gap. Some efficient training mechanism designed for Transformer to minimize the data amount requirement is also worth trying in the future \cite{liu2021efficient}. 

The introduction of global information is essential for 3D biomedical image segmentation. In this study by taking advantage of linear complexity self-attention, we introduced the long-distance modeling into the skip connection between encoder and decoder structure. We showed the feasibility of using linear self-attention for volumetric biomedical image segmentation and proposed one novel dynamic method to further reduce the computation burden. The promising result showed that the self-attention with linear complexity is worth more investigation in the near future for the biomedical image segmentation. In the future we will validate this model on different volumetric biomedical image segmentation including more organs and imaging modalities.

\subsubsection*{Acknowledgement}
This project is funded by the NIH grants: R01-CA246704 and R01-CA240639.

%
%
%
%
\bibliographystyle{splncs04}
\bibliography{mybibliography}
\end{document}